\documentclass[10pt,conference]{IEEEtran}
\IEEEoverridecommandlockouts
\usepackage{cite}
\usepackage{amsmath,amssymb,amsfonts}
\usepackage{algorithmic}
\usepackage{graphicx}
\usepackage{textcomp}
\usepackage{xcolor}
\usepackage{subcaption}

\usepackage[hidelinks,colorlinks,citecolor=gray,linkcolor=gray]{hyperref}
\usepackage{color,xcolor}
\usepackage{epsfig}
\usepackage{graphicx}

\usepackage{adjustbox}
\usepackage{array}
\usepackage{booktabs}
\usepackage{multirow}

\usepackage{amsmath,amsfonts,amsthm,amssymb}
\usepackage{bm}
\usepackage{nicefrac}
\usepackage{microtype}

\usepackage{changepage}
\usepackage{extramarks}
\usepackage{fancyhdr}
\usepackage{lastpage}
\usepackage{setspace}
\usepackage{soul}
\usepackage{xspace}

\def\BibTeX{{\rm B\kern-.05em{\sc i\kern-.025em b}\kern-.08em
    T\kern-.1667em\lower.7ex\hbox{E}\kern-.125emX}}
\begin{document}
	
\makeatletter
\newcommand{\linebreakand}{%
\end{@IEEEauthorhalign}
\hfill\mbox{}\par
\mbox{}\hfill\begin{@IEEEauthorhalign}
}
\makeatother

\title{Enabling On-Device Self-Supervised Contrastive Learning With Selective Data Contrast
\thanks{This work was supported in part by NSF CNS-2007274.}}


\author{
	\IEEEauthorblockN{Yawen Wu}
	\IEEEauthorblockA{\textit{Electrical and Computer Engineering} \\
		\textit{University of Pittsburgh}\\
		Pittsburgh, USA \\
		yawen.wu@pitt.edu}
	\and
	\IEEEauthorblockN{Zhepeng Wang}
	\IEEEauthorblockA{\textit{Electrical and Computer Engineering} \\
		\textit{University of Pittsburgh}\\
		Pittsburgh, USA \\
		zhepeng.wang@pitt.edu}
	\and
	\IEEEauthorblockN{Dewen Zeng}
	\IEEEauthorblockA{\textit{Computer Science and Engineering} \\
		\textit{University of Notre Dame}\\
		Notre Dame, USA \\
		dzeng2@nd.edu}
	\linebreakand 
	\IEEEauthorblockN{Yiyu Shi}
	\IEEEauthorblockA{\textit{Computer Science and Engineering} \\
		\textit{University of Notre Dame}\\
		Notre Dame, USA \\
		yshi4@nd.edu}
	\and
	\IEEEauthorblockN{Jingtong Hu}
	\IEEEauthorblockA{\textit{Electrical and Computer  Engineering} \\
		\textit{University of Pittsburgh}\\
		Pittsburgh, USA \\
		jthu@pitt.edu} 
}

\maketitle

\begin{abstract}
After a model is deployed on edge devices, it is desirable for these devices to learn from unlabeled data to continuously improve accuracy. Contrastive learning has demonstrated its great potential in learning from unlabeled data. However, the online input data are usually none independent and identically distributed (non-iid) and edge devices’ storages are usually too limited to store enough representative data from different data classes. We propose a framework to automatically select the most representative data from the unlabeled input stream, which only requires a small data buffer for dynamic learning. Experiments show that accuracy and learning speed are greatly improved.

\end{abstract}


\section{Introduction}

Deep learning models have been widely deployed on the edge and mobile devices to accomplish different tasks, such as robots 
for search and rescue \cite{shabbir2018survey} and UAVs for wildfire surveillance \cite{samaras2019deep}.
Traditionally, a model is pre-trained in high-performance servers and then deployed in these devices without further training \cite{jiang2019accuracy}. However, it is often desirable for these devices to learn from real-world input data (e.g. images captured by a camera) \cite{wu2020enabling} either based on a pre-trained model or totally from scratch when deployed to an unknown environment~\cite{pinto2016curious}. In this way, the model on robots or UAVs can adapt to new environments~\cite{she2020openloris}.

While it is feasible to send a few data to servers for labeling, it is prohibitive to send all these new data due to the requirement of expert knowledge, data privacy, communication cost, and latency concerns \cite{bonawitz2019towards}.
Thus, different from conventional training on servers by using fully labeled datasets, it is also desirable to learn from new streaming data in-situ with as few labels as possible. 

Contrastive learning, as an effective self-supervised learning approach \cite{ he2020momentum}, can learn visual representations from unlabeled data to improve the feature extractor (convolutional layers) in the model.
After contrastive learning, the classifier (fully connected layers) can be trained on top of the improved feature extractor by using few labeled data to achieve improved classification performance.
Contrastive learning is conventionally conducted by using a large dataset, which is completely collected before the training starts. 
In the learning process, 
each mini-batch is randomly sampled from the whole dataset to update the model \cite{chen2020simple}.
On edge platforms such as robots and UAVs, 
the data are collected by sensors such as cameras and continuously fed into the device. While it is theoretically possible to store the constantly generated massive unlabeled data on the device and employ contrastive learning, both the storage and energy overhead associated with writing and reading these data from storage devices (e.g. Flash memory) can be prohibitive in practice.

To learn from the unlabeled data stream without accumulating a large dataset,
a small data buffer can be used to form each mini-batch for training.
Existing contrastive learning frameworks \cite{chen2020simple, he2020momentum} assume that each mini-batch is independent and identically distributed (iid) by sampling uniformly at random from all the classes (i.e. each class has representative data in this mini-batch).
However, it is challenging to maintain the most representative data in the buffer such that learning from this buffer will efficiently reach an accurate model due to the following two reasons. \emph{First}, the streaming data collected on edge devices are usually temporally correlated \cite{orhan2020self} and result in a correlation within each mini-batch. 
This is because a long sequence of data in the temporally correlated stream can be in the same class \cite{hayes2019memory}.
For example, in wildlife monitoring,
goats from a group can appear in adjacent images captured by a continuous monitoring camera \cite{huynh2017deepmon} at some time, while zebras can appear in adjacent images at another time.
\emph{Second}, there is no easy way to select representative data for each class from the non-iid streaming data due to the fact that the streaming data are \emph{unlabeled}.
If labels were available for all the data, we could easily select representative data for each class \cite{hayes2019memory} based on all the labels even if the streaming data is non-iid.
Without addressing this challenge, directly learning from these temporally correlated non-iid mini-batches will result in slow learning speed and poor learned representations.

To improve the accuracy and expedite the learning process, it is essential to maintain a data buffer filled with representative data from the streaming data. 
To achieve this goal, this paper defines a contrast score, which is computed by the similarity between the features of a data and its flipped view. The contrast score of each data measures the quality of feature representation encoded by the model. Based on the contrast score, we propose a data replacement policy to maintain a representative data buffer. 
Data with a low quality of encoded representation by the model is more valuable for learning since they have not been effectively learned.
These data will be maintained in the buffer for further learning. 
On the other hand, data with a high quality of representations have been effectively learned, and they will be dropped to save places for more valuable data. After contrastive learning effectively learns from the unlabeled data and improve the feature extractor, the classifier needs to be updated as well. Since training the classifier without any labels does not generate meaningful accuracy, we will send as few as 1\% of the data to the server for labeling to improve the classifier and overall accuracy.

In summary, the main contributions of the paper include:
\begin{itemize}
    \item \textbf{Self-supervised on-device learning framework.} We propose a framework to form mini-batches of training data for self-supervised contrastive learning on-the-fly from the unlabeled input stream. It only uses a small data buffer and eliminates the necessity of storing all the streaming data into the device.
	\item \textbf{Contrast scoring for data selection.} We propose a data replacement policy by contrast scoring to maintain the most representative data in the buffer for on-device contrastive learning. 
    Labels are not needed in the data replacement process, and the selected data will generate large gradients that benefit the learning most.
	\item \textbf{Lazy scoring for reduced computation overhead.} We propose a lazy scoring strategy to reduce the runtime overhead of data scoring. The data scores are updated every several iterations instead of every iteration to save computation.
\end{itemize}

Experimental results on multiple datasets including CIFAR-10, CIFAR-100, SVHN, ImageNet-20, ImageNet-50, and ImageNet-100 show that the proposed framework achieves significantly higher accuracy than the state-of-the-art (SOTA) techniques and greatly improves the learning speed. 
With 1\% labeled data on the CIFAR-10 dataset, the proposed framework achieves 28.36\% higher accuracy than using the 1\% labeled data for direct supervised learning.
The proposed contrast scoring based data selection achieves 13.9\% higher accuracy than the SOTA data selection approach \cite{jiang2019accelerating}.
Meanwhile, the proposed approach achieves 2.67x faster learning than the baseline when the same accuracy is achieved.

\section{Background and Related Work}

\subsection{Background of Contrastive Learning}
Contrastive learning is a self-supervised approach to learn an encoder (feature extractor) for extracting visual representations from the input image.
In this work, we employ the contrastive learning approach from \cite{chen2020simple} since it performs on par with its supervised counterpart.
For an input image $x$, its representation vector $h$ is obtained by $h=f(x)$, 
where $f(\cdot)$ is the backbone of a deep learning model (i.e. convolutional layers).
To boost the performance of learned representation, a project head $g(\cdot)$ is used to map the data representation to the latent space as a vector $z=g(h)=g(f(x))$ where contrastive loss is applied. 
To create a positive pair $(z_i, z_{i^{+}})$, one input $x$ is augmented twice as $(x_i, x_{i^{+}})$ and then fed into the encoder to get representation vectors $(h_i,h_{i^{+}})=(f(x_i),f(x_{i^{+}}))$, which are further projected by $g(\cdot)$ and normalized as 
$(z_i,z_{i^{+}})$.
Then for each positive pair $(z_i, z_{i^{+}})$ in one mini-batch,
the contrastive loss 
is applied
to compute the loss $\ell_{i, i^{+}}$ as follows:
\setlength{\abovedisplayskip}{0pt}
\setlength{\belowdisplayskip}{0pt}
\setlength{\abovedisplayshortskip}{0pt}
\setlength{\belowdisplayshortskip}{0pt}
\begin{equation}\label{equ:contrastloss}
    \ell_{i, i^{+}}=-\log \frac{\exp (z_{i} \cdot z_{i^{+}} / \tau)}{\exp (z_{i} \cdot z_{i^{+}} / \tau) + \sum_{i^{-}} \exp (z_{i} \cdot z_{i^{-}} / \tau)}
\end{equation}
where $z_{i^{-}}$ is the representation vector of other data (serving as negatives to contrast with) in the same mini-batch, and $\tau$ is the temperature.
Minimizing $\sum \ell_{i, i^{+}}$ in one mini-batch by iteratively updating the model will learn an encoder to generate representations.

\subsection{Related Work}
\textbf{Contrastive Visual Representation Learning.}
\cite{he2020momentum, chen2020simple} employ contrastive loss for representation learning and achieve high accuracy on classification and segmentation tasks.
\cite{knights2020temporally,orhan2020self} use the temporal correlations in the streaming data to improve representation learning.
However, all these works assume that the whole training dataset is available in the learning process, and each mini-batch can be formed by sampling from the dataset. 
Each mini-batch consists of independent and identically distributed (iid) data.
But when learning from the streaming data, which cannot be assumed to be iid on edge devices, the data is collected sequentially as it is.
Besides, random sampling from the entire input stream to create iid mini-batches is infeasible since it requires storing all the data. Therefore, an approach to form mini-batches on-the-fly while including the most representative data in each mini-batch is needed to enable efficient and accurate on-device contrastive learning.

\textbf{Data Selection in Streaming and Continual Learning.}
There are several supervised streaming and continual learning models that can learn from a stream of data \cite{aljundi2019gradient}.
To overcome the problem of catastrophic forgetting of previously seen data,
a data buffer is usually needed to store previous data for rehearsal \cite{hayes2019memory, aljundi2019gradient, borsos2020coresets}. 
The main drawback of these approaches is that \emph{data labels} are needed to maintain the buffer.
However, labeling all the data in the streaming is prohibitive or even infeasible on edge devices.
Therefore, existing methods cannot be applied directly to contrastive learning and an effective data selection approach that works on \emph{unlabeled} data is needed.

\section{Self-Supervised On-Device Learning by Selective Data Contrast}

This paper proposes a framework to efficiently learn data representations from the unlabeled input stream on-the-fly without accumulating a large dataset due to storage limitations on edge devices.
To maintain the most representative data in the buffer such that learning from these data will benefit the model most, we propose a data replacement policy based on \emph{Contrast Score} by measuring the quality of representation for each data without using labels. 
Data with low quality of representations have not been effectively learned by the model and will be maintained in the buffer for further learning, while data with high quality of representations will be dropped.
The contrast scoring is supported by the theoretical analysis that data with higher scores will generate larger gradients and accelerate the learning process.

In this section, we will first present the framework overview in Section \ref{sec:overview}. Then we will introduce the proposed contrast scoring for data selection in Section \ref{sec:contrast_score}. 
After that, we will theoretically analyze 
the effectiveness of contrast scoring in Section \ref{sec:contrast_grad}.
Finally, we will introduce lazy scoring to reduce the runtime overhead of contrast scoring in Section \ref{sec:lazy_score}.

\subsection{Framework Overview}\label{sec:overview}
\begin{figure}[!htb]
	\centering
	\vspace{-12pt}
	\includegraphics[width=\linewidth]{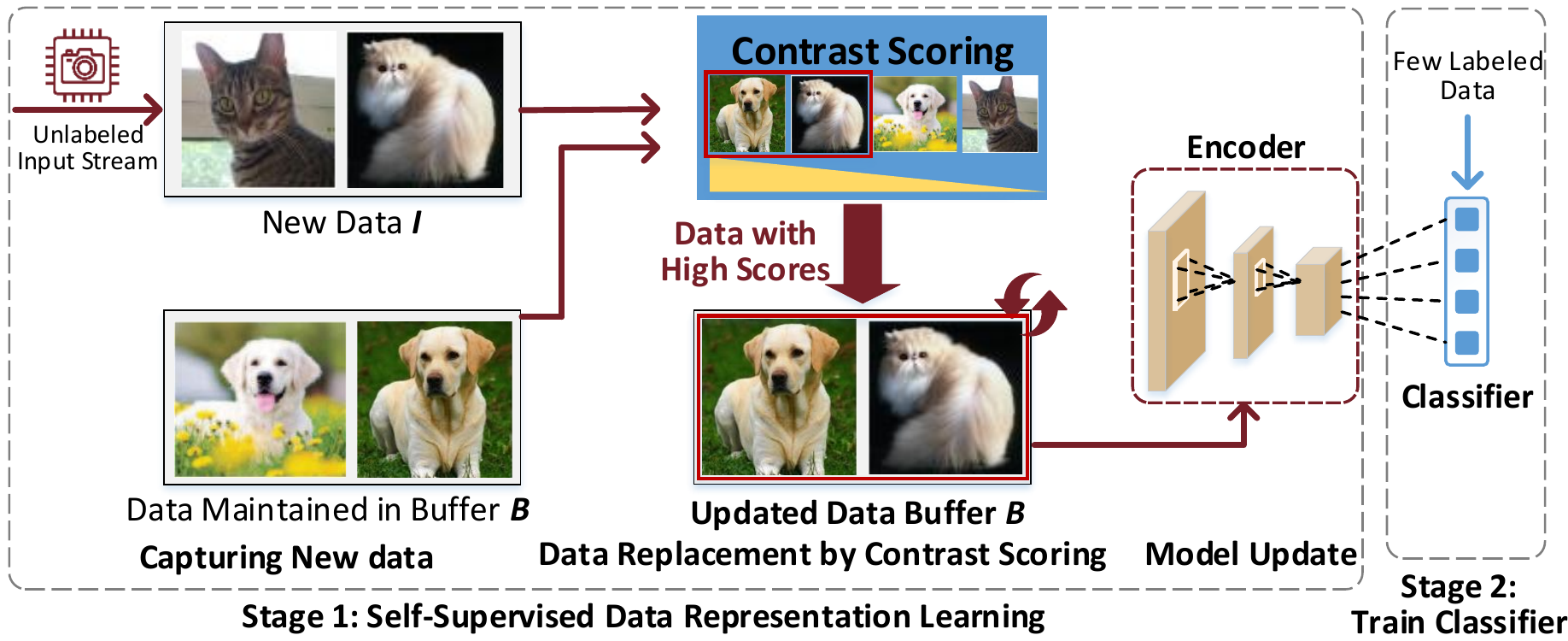}
	\caption{Overview of on-device contrastive learning framework. The encoder is first trained by contrastive learning with data selected from unlabeled streaming data by contrast scoring, and then the classifier is trained by few (e.g. 1\%) labeled data.}
	\vspace{-10pt}
	\label{fig:overview}
\end{figure}

As shown in Fig. \ref{fig:overview}, the proposed framework has two stages. The first stage learns an encoder (i.e. convolutional layers) by self-supervised contrastive learning to generate data representations (i.e. low-dimensional vectors) from the high-dimensional unlabeled inputs (e.g. images). The second stage learns a classifier by using few (e.g. 1\%) labeled data on top of the learned representations. 

In stage 1, the proposed framework consumes the input streaming data on-the-fly to update the model for improved representation. We only use a small data buffer $B$ (i.e. the same size as one mini-batch) to maintain the most representative data. 
When a segment of new input $I$ arrives, both the new data in $I$ and the data in the buffer $B$ will be scored to find the most representative data. 
While any size of $I$ can be used, for simplicity we assume $I$ has the size as $B$ by setting $\text{size}(I)= \text{size}(B)$.
Then the data with the highest scores in $B \cup I$ will be selected and put into $B$.
In this way, the data replacement process always maintains the most representative data among the new and the old ones. 
After each iteration of data replacement, the data preserved in the data buffer $B$ will serve as one mini-batch for updating the model once.
The detailed data replacement policy will be described in the next subsection.

\subsection{Data Replacement By Contrast Scoring}\label{sec:contrast_score}

\begin{figure}[!htb]
	\centering
	\includegraphics[width=1.0\linewidth]{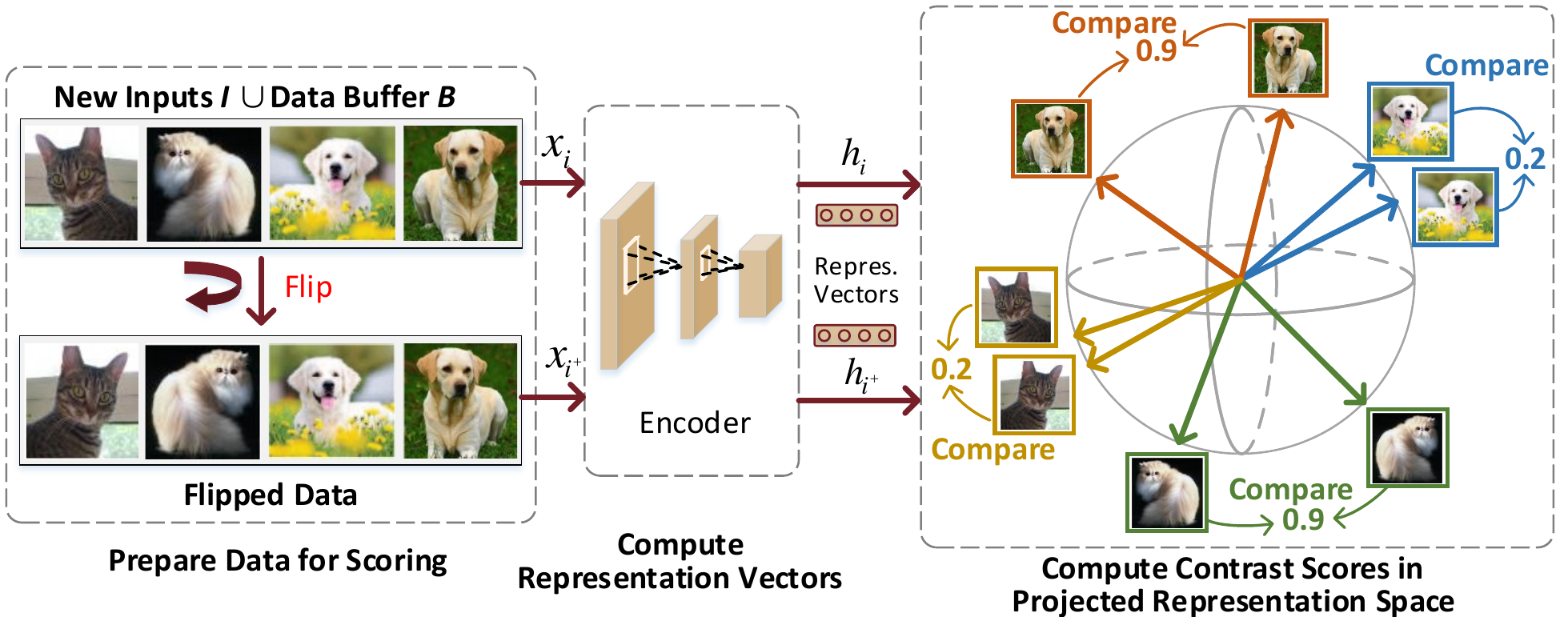}
	\vspace{-16pt}
	\caption{Contrast scoring for data replacement. The original and flipped inputs are fed into the encoder to generate representation vectors, which are projected to vectors in the unit sphere to compute scores.}
	\vspace{-6pt}
	\label{fig:scoring}
\end{figure}

\textbf{Contrast Scoring.}
For each input $x_i$, the \emph{contrast scoring} function $S(x_i)$ aims to measure the quality of the representation vector $h_i=f(x_i)$ generated by the base encoder $f(\cdot)$. 
Intuitively, if the representation of $x_i$ is not good, $x_i$ will be valuable data for updating the base encoder since it can still learn from $x_i$ to improve its capability of encoding $x_i$.
To achieve this, as shown in Fig. \ref{fig:scoring}, for each image $x_i$ from the input stream and the buffer, we generate another view $x_{i^+}$ by horizontal flipping. Then we feed both $x_i$ and $x_{i^+}$ into the encoder and generate the representation vectors $h_i$ and $h_{i^+}$ for these two views. 
Ideally, if the encoder has learned to generate effective representations of $x_i$, $h_i$ and $h_{i^+}$ will be identical or very similar.
After that, based on $h_i$ and $h_{i^+}$, the score for $x_i$ is computed by the contrast scoring function $S(\cdot)$.

The contrast scoring function $S(\cdot)$ is defined as: 
\begin{equation} \label{equ:score}
\begin{split}
S(x_i)=dissim(x_i,x_{i^+})&=1-similarity(z_i, z_{i^+}) \\
 & = 1-z_i^T z_{i^+},\quad x_i \in \{ {B} \cup {I} \}\\
\end{split}
\end{equation}
\begin{equation}
\text {where } z_i=g(h_i) / \|g(h_i)\|_{\ell_2},\  z_{i^+}=g(h_{i^+}) / \|g(h_{i^+})\|_{\ell_2}
\end{equation}
where $h_i$ and $h_{i^+}$ are the representation vectors generated by the base encoder $f(\cdot)$ as $h_i=f(x_i)$ and $h_{i^+}=f(x_{i^+})$, taking data $x_i$ and its horizontally flipped view $x_{i^+}$ as inputs, respectively.
$z_i$ and $z_{i^+}$ are $\ell_2$-normalized vectors from the projection head $g(\cdot)$ to enforce $\|z_i\|_{\ell_2}=\|z_{i^+}\|_{\ell_2}=1$. In this way, the dot product $z_i^T z_{i^+}$ is in the range [-1,1], and $S(x_i)$ is non-negative and in the range [0,2].

The contrast scoring function Eq.(\ref{equ:score}) measures the dissimilarity between the projected representation vectors of an image $x_i$ and its horizontal flip $x_{i^+}$, where \textit{a higher score means a larger dissimilarity}. 
Essentially, the representation of one image needs to be invariant to image transformations \cite{ji2019invariant}, and the representations of $x_i$ and $x_{i^+}$ need to be as similar as possible. Since a higher score represents a larger dissimilarity and less invariance, input $x_i$ with a higher score is more valuable for updating the base encoder because the base encoder still cannot generate sufficiently good representations of it.
By updating the encoder with $x_i$ using the contrastive loss \cite{chen2020simple}, which aims to maximize the similarity of two strongly augmented views of $x_i$, the score of $x_i$ in Eq.(\ref{equ:score}) will decrease and $x_i$ will have a lower probability of being selected into the next mini-batch in Eq.(\ref{equ:nextbatch}).
In this way, more valuable data to update the base encoder will have a higher probability of being selected into the next mini-batch and others are more likely to be dropped. 
A detailed analysis of the effectiveness of contrast scoring will be provided in Section \ref{sec:contrast_grad}.

\textbf{Contrast Score Design Principle.}
Contrast scoring is a metric to represent the capability of the base encoder in generating the representation $h_i=f(x_i)$ for $x_i$. Thus, it should only relate to the image itself and the encoder. In Fig. \ref{fig:scoring}, when generating a pair of inputs $(x_i,x_{i^+})$ to $S(\cdot)$ from an image $x_i$, we find it crucial to avoid any randomness (e.g. random crop) and \textit{only apply the weak data augmentation} (i.e. horizontal flipping) to generate $x_{i^+}$.
The reason is that this weak augmentation is deterministic and provides consistent inputs to $S(\cdot)$.
In this way, the score $S(\cdot)$ is deterministic to $x_i$ and is consistent in different runs of $S(\cdot)$. If strong data augmentation such as random crop and random color distortion were used here, the score will change on the same input in different runs of $S(\cdot)$ due to the randomness introduced in these augmentation techniques. Involving the randomness into the inputs to the contrast scoring function will generate biased scores, in which the scores will mainly reflect the randomness. In that case, the score will not be an objective evaluation of the encoder's capability. 

\textbf{Contrast Score Based Data Selection.}
At iteration $t$, the goal is to form the next mini-batch ${B}_{t+1}$ by selecting the most informative data from ${I}_t$ and ${B}_t$, such that learning from ${B}_{t+1}$ will benefit the model most.
To achieve this, we apply the contrast scoring function $S(\cdot)$ to both the data already in the buffer $B_t$ and new data $I_t$.
$B_{t+1}$ is formed by selecting the data with the highest contrast scores in ${B}_{t} \cup {I}_{t}$:
\begin{equation}\label{equ:nextbatch}
{B}_{t+1}=\{x_i | x_i \in {B}_{t} \cup {I}_{t}, i \in topN(\{S(x_i)\}_{i=1}^{2N}) \}
\end{equation}
where $topN()$ returns the indices of $x_i$ with the top $N$ scores.
In this way, the most representative data is maintained in the buffer by using the proposed contrast scoring.

\subsection{Effectiveness of Contrast Score}\label{sec:contrast_grad}
The proposed contrast scoring effectively selects data that can generate large gradients, which benefits the learning most.
To understand this, for each data $x_i$ in one mini-batch, the gradient of contrastive loss $\ell_{i, i^{+}}$ in Eq.(\ref{equ:contrastloss}) with respect to the representation vector $z_i$ is computed as:
\begin{equation}\label{equ:contrast_grad}
\frac{\partial \ell_{i, i^{+}}}{\partial {z_i}}=-\frac{1}{\tau}\left(\left(1-p_{z_{i^+}}\right) \cdot {z_i}-\sum_{z_{i^-}} p_{z_{i^-}} \cdot {z_{i^-}}\right),
\end{equation}
\begin{equation}\label{equ:contrast_prob}
\text {where } p_{z}=\frac{\exp \left({z_i}^{T} {z} / \tau\right)}{\sum_{z_j \in \{z_{i^+},z_{i^-}\}} \exp \left({z_i}^{T} {z}_{j} / \tau\right)},\quad p_{z} \in \{p_{z_{i^+}}, p_{z_{i^-}}\}
\end{equation}
$p_z$ is the probability distribution generated by applying the softmax function to the similarity ${z_i}^T z_j$ between $z_i$ and each 
$\ z_j \in \{z_{i^+},z_{i^-}\}$ in the mini-batch.
For $z=z_{i^+}$, $p_{z_{i^+}}$ is the matching probability of $z_i$ with its positive pair $z_{i^{+}}$.
Similarly, for $z=z_{i^-}$, $p_{z_{i^-}}$ is the matching probability of $z_i$ with a negative pair $z_{i^{-}}$ (i.e. the representation vector of other data in the same mini-batch).

A data with a small contrast score $S(x_i)$ generates a near-zero gradient and contributes almost nothing to the learning process.
On the other hand, a data $x_i$ with a high contrast score $S(x_i)$ in Eq.(\ref{equ:score}) corresponds to a large gradient in Eq.(\ref{equ:contrast_grad}), which contributes much to the learning process.
To understand this, we analyze the relationship between the contrast score in Eq.(\ref{equ:score}) and the gradient in Eq.(\ref{equ:contrast_grad}) in two cases:

\textbf{Case 1: A data with a small contrast score generates a near-zero gradient.}
A small contrast score in Eq.(\ref{equ:score}) corresponds to a large similarity between ${z_i}$ and $z_{i^+}$.
Therefore, the value of dot product ${z_i}^T z_{i^+}$ will be large and dominate the elements in the softmax function in Eq.(\ref{equ:contrast_prob}). As a result, $p_{z_{i^+}}$ will be large and near 1. 
Since $p_{z_i^{+}}+\sum_{z \in z_{i^-}} {p_z} =1$ as a property of the softmax function, the values of all $p_{z_{i-}}$ will be small and near 0.
In this way, $1-p_{z_{i^+}}$ as well as all $p_{z_{i^-}}$ will be near 0, and the gradient $\frac{\partial \ell_{i, i^{+}}}{\partial {z_i}}$ will be near 0.
Using the near-zero gradient to perform one gradient descent step $w \leftarrow w - \eta \frac{\partial \ell_{i, i^{+}}}{\partial {z_i}}$
does not contribute to the learning since there is almost no change in the weight.

\textbf{Case 2: A data with a high contrast score generates a large gradient.}
When the contrast score is high, ${z_i}$ and $z_{i^+}$ are dissimilar to each other. By applying the same reasoning as case 1, $1-p_{z_{i^+}}$ and all $p_{z_{i^-}}$ will be near 1, and the gradient $\frac{\partial \ell_{i, i^{+}}}{\partial {z_i}}$ will be large, which significantly contributes to the learning process.

Therefore, by using the proposed contrast score, 
trivial data that only generate near-zero gradients will be dropped while
important data that can generate large gradients will be maintained in the buffer for learning.

\subsection{Lazy Scoring}\label{sec:lazy_score}
Computing the scores for new data and data in the buffer requires feeding these data into the base encoder to generate the representations. This computation incurs additional time overhead. To minimize the overhead, we propose lazy scoring, in which part of the data scores can be reused to reduce computation.

We made the following two observations as the foundation of lazy scoring. 
\emph{First}, during each iteration of data replacement, most of the data (i.e. about 90\%) in the buffer are preserved while most of the new data are directly dropped.
Therefore, by reusing contrast scores of data in the buffer, a large portion of the computation in scoring can be reduced.
\emph{Second}, the score $S(x_i)$ of data $x_i$ only slightly changes across several adjacent iterations. This is because the score of data $x_i$ only depends on itself and the base encoder $f(\cdot)$. $x_i$ remains constant and $f(\cdot)$ is slowly updated across iterations. Therefore, $S(x_i)$ is only slowly updated following the pace of $f(\cdot)$, and the score $S(x_i)$ computed iterations ago still provides meaningful information of $x_i$. 

To achieve lazy scoring, as long as data $x_i$ remains in buffer $B$, its score is updated every $T$ iterations instead of in every iteration. 
More specifically, for each $x_i$ in ${B}$, we track its age $age(x_i)$ in the number of iterations since it was placed in $B$. 
When performing scoring, we separate data in ${B}$ into two subsets, in which one needs scoring while the other does not. The subset of data that needs scoring is denoted as:
\begin{equation}\label{eq:lazy_score_set}
{B}_t^{\prime} =\{x_i\ |\ x \in {B}_t \ \text{and}\ age(x_i)\ mod\ T=0\}
\end{equation}
When scoring data in $B$, the scores are updated as:
\begin{equation}\label{equ:}
S_{t}(x_i) = 
\begin{cases}
dissim(x_i, x_{i^+}), &\quad x_i \in {B}_t^{\prime}\\
S_{t-1}(x_i), & \quad \text{otherwise}
\end{cases}
\end{equation}
In the above equation, if $x_i$ needs scoring, its score is computed by Eq.(\ref{equ:score}). Otherwise, its score in the last iteration is copied to save computation.
By lazying scoring, the computation overhead of contrast scoring is effectively reduced to about $\frac{1}{T}$ of that without lazy scoring, while the accuracy is preserved.

\section{Experiments}
In this section, we first evaluate the \emph{accuracy} with different labeling ratios. Then, we evaluate the \emph{learning speed} of the proposed framework. After that, 
we evaluate the \emph{reduced computation overhead} by lazy scoring. 
Finally, we evaluate the impact of \emph{buffer size}.

\subsection{Experimental Setup}

\textbf{Datasets and Evaluation Protocols.}
We use multiple datasets, including CIFAR-10, CIFAR-100 \cite{krizhevsky2009learning}, SVHN \cite{netzer2011reading}, ImageNet-20/50/100 \cite{russakovsky2015imagenet} to evaluate the proposed approaches, which are widely used for edge devices \cite{jiang2020hardware,wu2020intermittent}.
To perform classification, the encoder is first trained by the proposed approaches to generate data representations.
As we mentioned before, training a classifier without any labels does not generate meaningful accuracy. Therefore, we train a classifier with 1\%, 10\%, or 100\% labeled data on the learned encoder.

\textbf{Strength of Temporal Correlation (STC).}
We use the metric Strength of Temporal Correlation (STC) to represent the temporal correlation of the input stream.
STC represents how many consecutive data in the input stream are from the same class  until a class change happens \cite{hayes2019memory}. A larger STC represents a stronger temporal correlation.

\textbf{Default Training Setting.} \label{sec:app_training_setting}
We use ResNet-18 as the base encoder.
We train the encoder with the contrastive loss \cite{chen2020simple} by the Adam optimizer.
While the proposed approaches can be applied to both training \emph{from scratch} and \emph{fine-tuning} a pre-trained model, to avoid any bias in the pre-trained model on any approach to compare with, we train \emph{from scratch}.
Unless otherwise specified, the batch size is 256 with the weight decay 0.0001.
For subsets of ImageNet, the learning rate is 0.0004, the temperature $\tau$ is 0.07, and STC is 100. The model is trained for 300 epochs for ImageNet-20/50 and 100 epochs for ImageNet-100.
For CIFAR-10, CIFAR-100, and SVHN, the learning rate is 0.0001, the temperature $\tau$ is 0.5, and the model is trained for 500 epochs with STC 500.
For all datasets, the classifier is trained for 500 epochs with Adam optimizer and learning rate 0.0003.
The lazy scoring is disabled by default to have a fair comparison of different data replacement approaches.
The results are averaged over three runs on 2 Nvidia V100 GPUs with different random seeds.

\textbf{Baselines.}
We first compare the proposed framework with supervised learning using 1\% or 10\% labeled data. Then, we compare the proposed contrast scoring with four data selection baselines which select data from \emph{unlabeled streaming}.
The first two baselines are
popular and effective strategies for maintaining exemplars in continual learning while not requiring labels.
\textit{Random replacement} is a variant of reservoir sampling \cite{vitter1985random} and is recently used for continual learning \cite{hayes2019memory}. It selects data uniformly at random from new data and data in the buffer to form the new data buffer.
\textit{FIFO replacement} is also recently employed for continual learning \cite{hayes2019memory}. It replaces the oldest data in the buffer with new data.
While not requiring labeling information and seemingly simple, these two approaches have demonstrated superior performance in maintaining data for continual learning compared with approaches that rely on exact labels \cite{chaudhry2019continual}.
The next two baselines are SOTA approaches to select data for efficient training and improving accuracy.
\emph{Selective-Backprop} \cite{jiang2019accelerating} selects data with the largest losses for training.
\emph{K-Center} is a SOTA active learning approach \cite{sener2017active}, which selects the most representative data by performing k-center clustering in the features space.
For conciseness, in the following figures and tables, we use \textbf{\emph{Contrast Scoring}} to represent the proposed approaches, and use \textbf{\emph{Random Replace}}, \textbf{\emph{FIFO Replace}}, \textbf{\emph{Selective-BP}}, and \textbf{\emph{K-Center}} to represent the baselines.

\subsection{Improved Accuracy with Different Labeling Ratios}

\begin{figure}[!htb]
	\centering
	\vspace{-12pt}
	\includegraphics[width=1.0\linewidth]{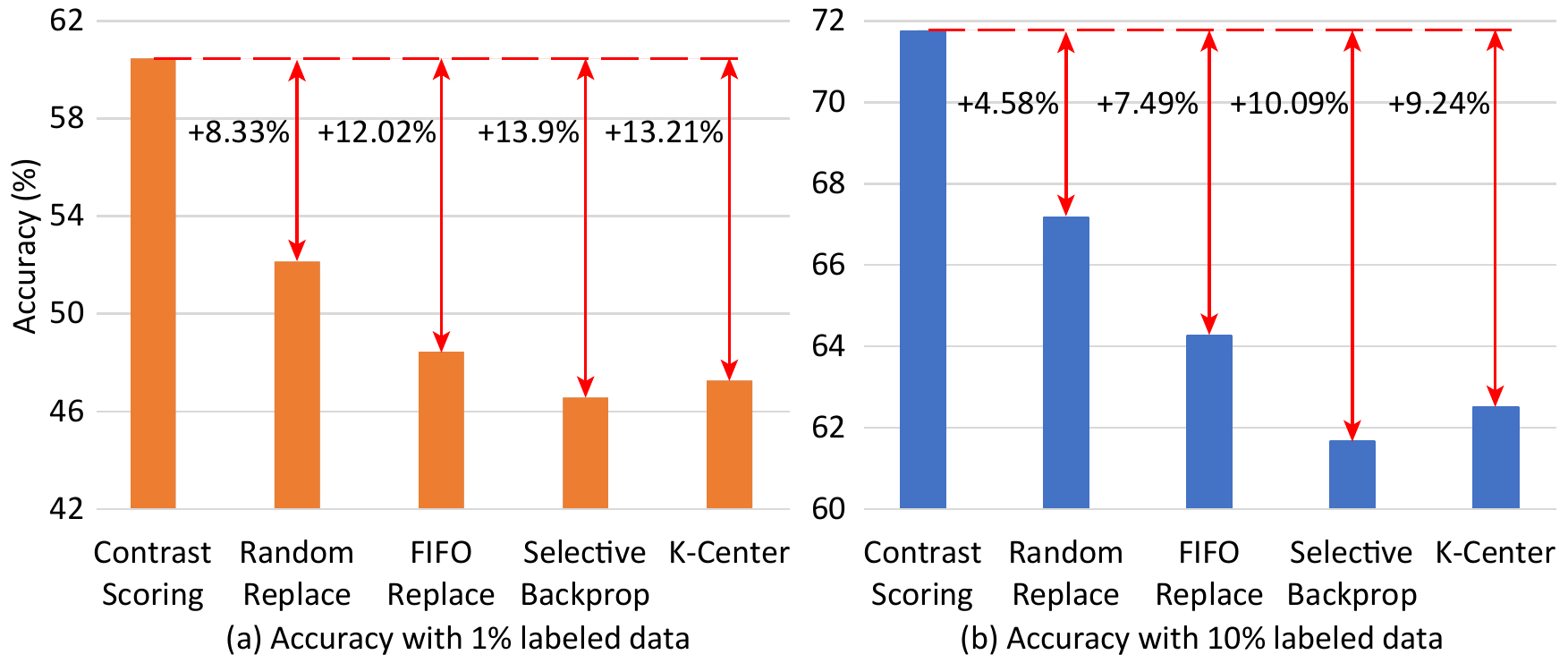}
	\vspace{-16pt}
	\caption{Accuracy on CIFAR-10 with 1\% and 10\% labeled data.}
	\vspace{-10pt}
	\label{fig:exp_multiple_baselines}
\end{figure}

We first compare the proposed framework with supervised learning using 1\% or 10\% labeled data.
The supervised learning achieves the accuracy of 32.11\% and 40.53\%, which are 28.36\% and 31.22\% lower than the proposed approaches. Therefore, supervised learning is not a practical option, and we will focus on evaluating the accuracy of the proposed framework with different data selection approaches.

We compare the proposed contrast scoring with other data selection approaches in terms of accuracy 
by first performing contrastive learning on unlabeled data with different approaches, and then learning the classifier with different ratios of labeled data (i.e. 1\%, 10\%).

The proposed data selection approach by contrast scoring substantially outperforms the SOTA baselines.
The accuracy with different labeling ratios (i.e. 1\%, 10\%) on CIFAR-10 is shown in Fig. \ref{fig:exp_multiple_baselines}, in which the contrastive learning is performed for 100 epochs without labels before training the classifier.
First, with 1\% and 10\% labeled data for learning the classifier, the proposed Contrast Scoring achieves the accuracy of 60.47\% and 71.75\%, and outperforms other four approaches by \{8.33\%, 12.02\%, 13.9\%, 13.21\%\} and \{4.58\%, 7.49\%, 10.09\%, 9.24\%\}, respectively.
Second, 
with fewer labels (i.e. 1\% vs. 10\%),
the proposed contrast scoring outperforms each baseline by a larger margin. This is because with fewer labels, the quality of learned representation becomes more important, and the proposed framework learns better representations than the baselines.
Third, the SOTA approaches \emph{Selective-BP} and \emph{K-Center} are designed to select data for training with supervised cross-entropy loss, and the data selected by them do not benefit self-supervised contrastive learning. 
Different from this, the proposed Contrast Scoring selects data that benefit contrastive learning the most.

The results show that the most competitive baselines are the two seemingly simple, yet surprisingly effective approaches \emph{Random Replace} and \emph{FIFO Replace}.
These results match the results in \cite{borsos2020coresets}, where a random replacement policy outperforms elaborately designed approaches.

\subsection{Learning Curve: Improved Learning Speed and Accuracy}
We evaluate the learning curve of the proposed approaches and baselines on CIFAR-10, ImageNet-20, ImageNet-50, ImageNet-100, SVHN, and CIFAR-100 datasets. 
The learning curve represents how fast the model learns representations from the new inputs.
Since we aim to evaluate the contrastive learning process by different data selection approaches,
to avoid the influence of different label ratios in training the classifier,
in the following evaluations, we will use 100\% labeled data to train the classifier after contrastive learning and only compare with the two most competitive baselines.

\begin{figure}[!htb]
	\centering
	\vspace{-16pt}
	\begin{subfigure}[b]{0.45\linewidth}
		\includegraphics[width=\linewidth]{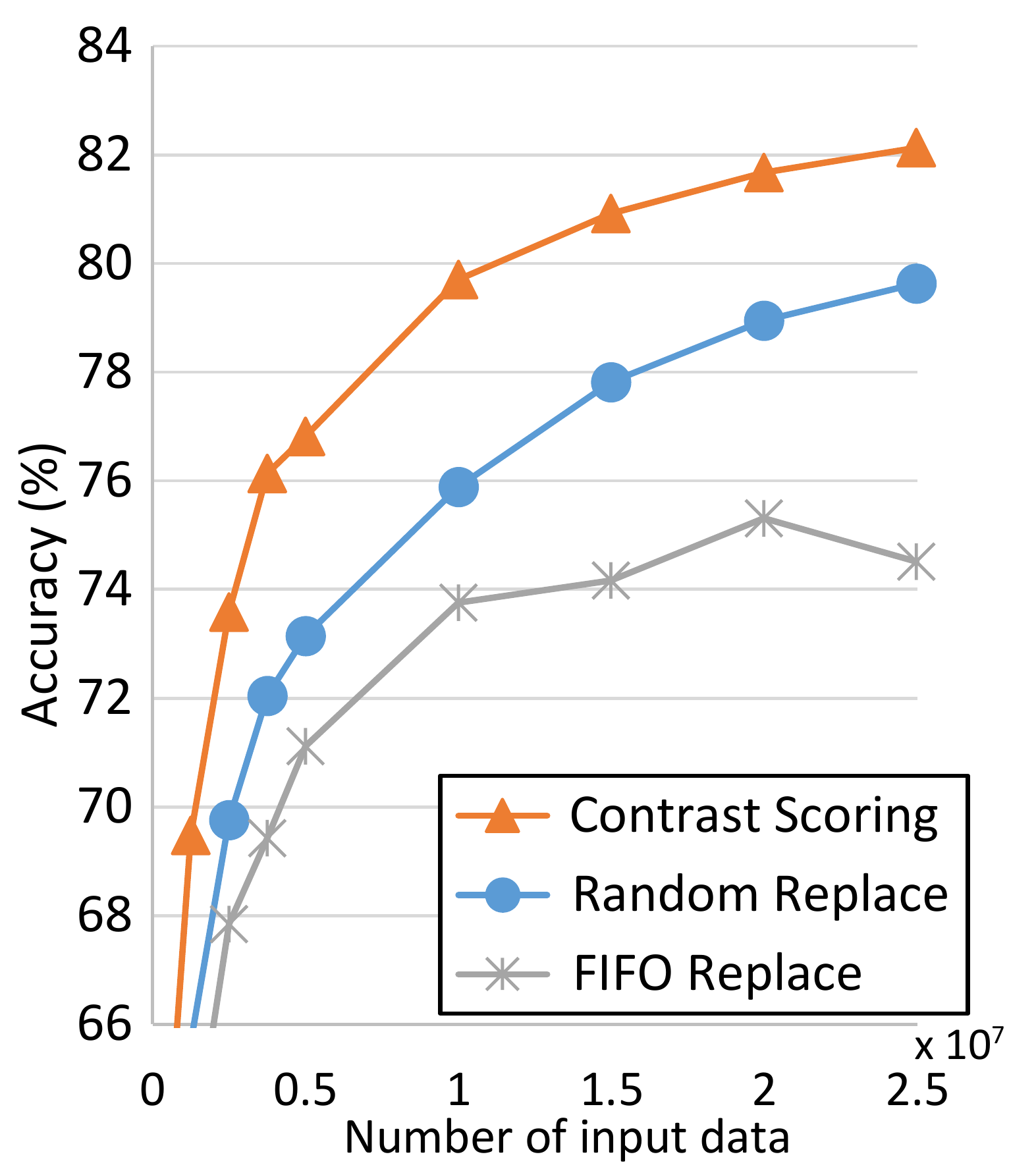}
		\vspace{-16pt}
		\caption{CIFAR-10.}
		\label{fig:}
	\end{subfigure}
	\begin{subfigure}[b]{0.45\linewidth}
		\includegraphics[width=\linewidth]{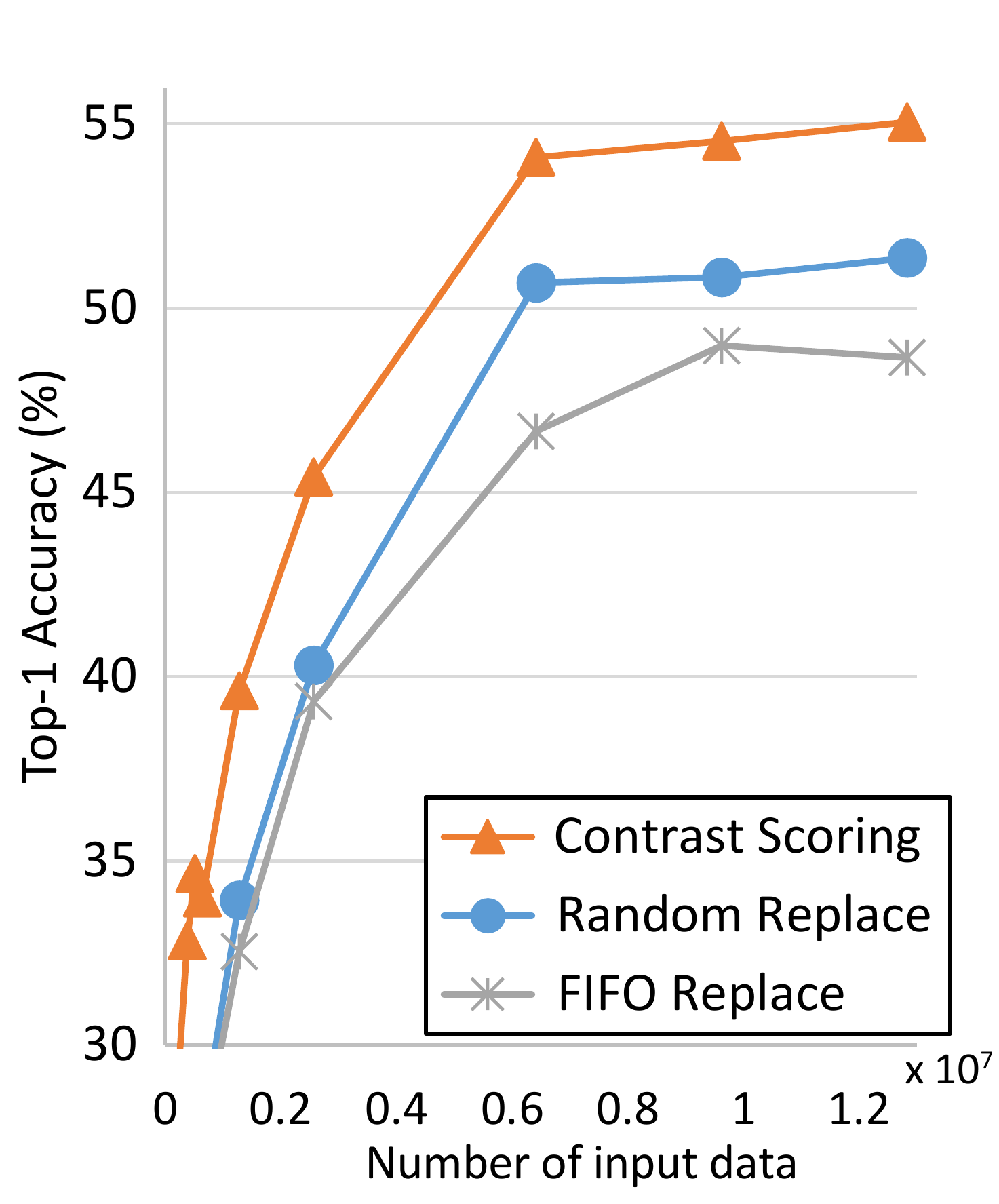}
		\vspace{-15pt}
		\caption{ImageNet-100.}
		\label{fig:}
	\end{subfigure}
	\vspace{-6pt}
	\caption{Learning curve on CIFAR-10 and ImageNet-100 datasets. 
	The learned representations by the proposed data replacement with scoring substantially outperform the baselines under two evaluation protocols.}
	\label{fig:learning_curve_cifar10}
	\vspace{-8pt}
\end{figure}

\textbf{Learning Curve on CIFAR-10.}
The proposed data replacement policy quickly learns data representations and achieves a significantly faster learning speed and a higher accuracy than the baselines. 
The learning curve on CIFAR-10 is shown in Fig. \ref{fig:learning_curve_cifar10} (a). The $x$-axis is the number of seen inputs and the $y$-axis is the accuracy.
The accuracy of the proposed approaches quickly increases to 76.1\% with 3.74M seen data, which is 2.67$\times$ faster than  the random replacement policy that needs 9.98M data to achieve similar accuracy. The FIFO replacement policy cannot achieve this accuracy even with 25M data. 
Besides, the proposed approaches achieve a much higher final accuracy than the baselines. The proposed approaches achieve a final accuracy of 82.13\%, while the random and FIFO replacement policies only achieve 79.63\% and 74.51\%, respectively.

\textbf{Learning Curve on ImageNet-100.}
We further evaluate the proposed approaches on the ImageNet-100 dataset.
While this dataset is a subset of the large-scale ImageNet dataset, it still features high-resolution images and is challenging for the stream setting.
As shown in Fig. \ref{fig:learning_curve_cifar10} (b), the proposed approaches achieve a consistently faster learning speed than the baselines. 
The proposed approaches achieve 55.05\% top-1 accuracy and outperform the baselines by 3.69\% and 6.39\%, respectively.

\begin{figure}[ht]
	\centering
	\vspace{-10pt}
	\begin{subfigure}{0.45\linewidth}
		\includegraphics[width=\linewidth]{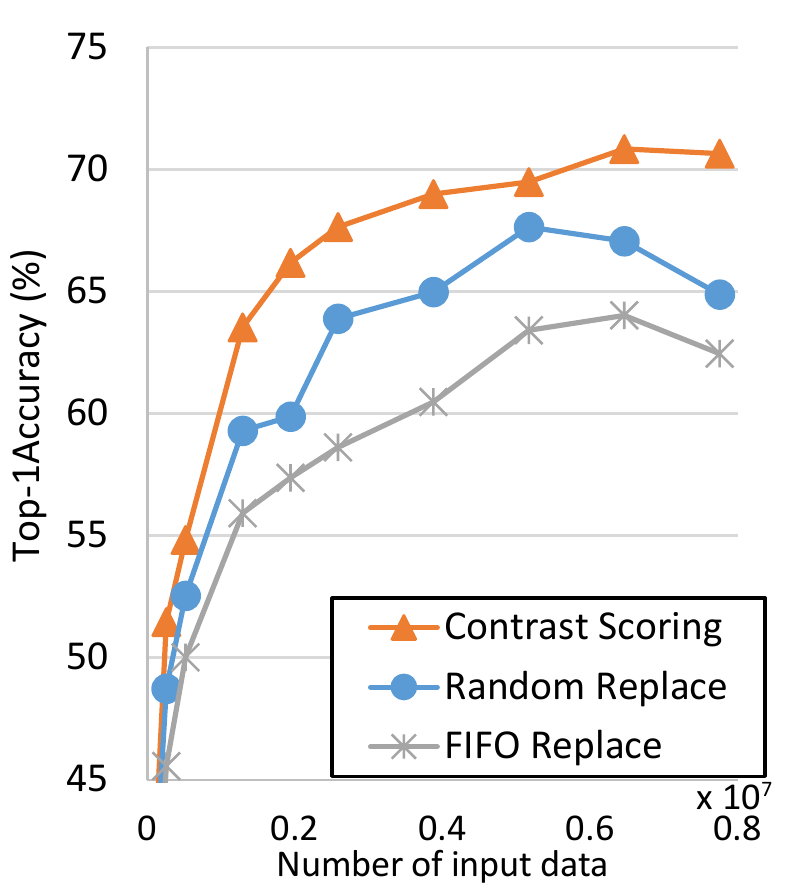}
		\vspace{-15pt}
		\caption{ImageNet-20.}
		\label{fig:}
	\end{subfigure}
	\begin{subfigure}{0.45\linewidth}
		\includegraphics[width=\linewidth]{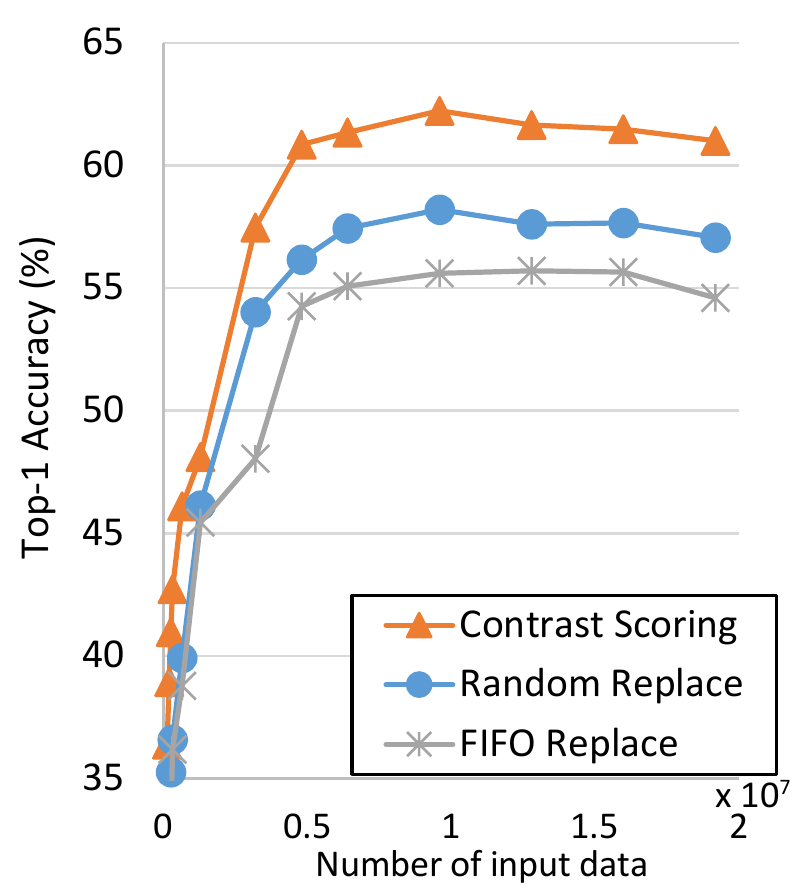}
		\vspace{-15pt}
		\caption{ImageNet-50.}
		\label{fig:}
	\end{subfigure}
	\vspace{-6pt}
	\caption{Learning curve on ImageNet-20 and ImageNet-50 dataset.}
	\label{fig:learning_curve_imagenet20}
	\vspace{-8pt}
\end{figure}

\textbf{Learning Curve on ImageNet-20 and ImageNet-50.}
We evaluate the proposed approaches on the ImageNet-20 and ImageNet-50 dataset. 
As shown in Fig. \ref{fig:learning_curve_imagenet20}, 
the proposed approaches achieve a significantly faster learning speed and higher accuracy than the baselines.
On ImageNet-20, the proposed approaches achieve 70.64\% top-1 accuracy and outperform two baselines by 5.76\% and 8.19\%, respectively.
On ImageNet-50, the proposed approaches achieve 60.99\% top-1 accuracy and outperform the baselines by 3.94\% and 6.39\%, respectively.

\begin{figure}[!htb]
	\centering
	\vspace{-16pt}
	\begin{subfigure}[b]{0.45\linewidth}
		\includegraphics[width=\linewidth]{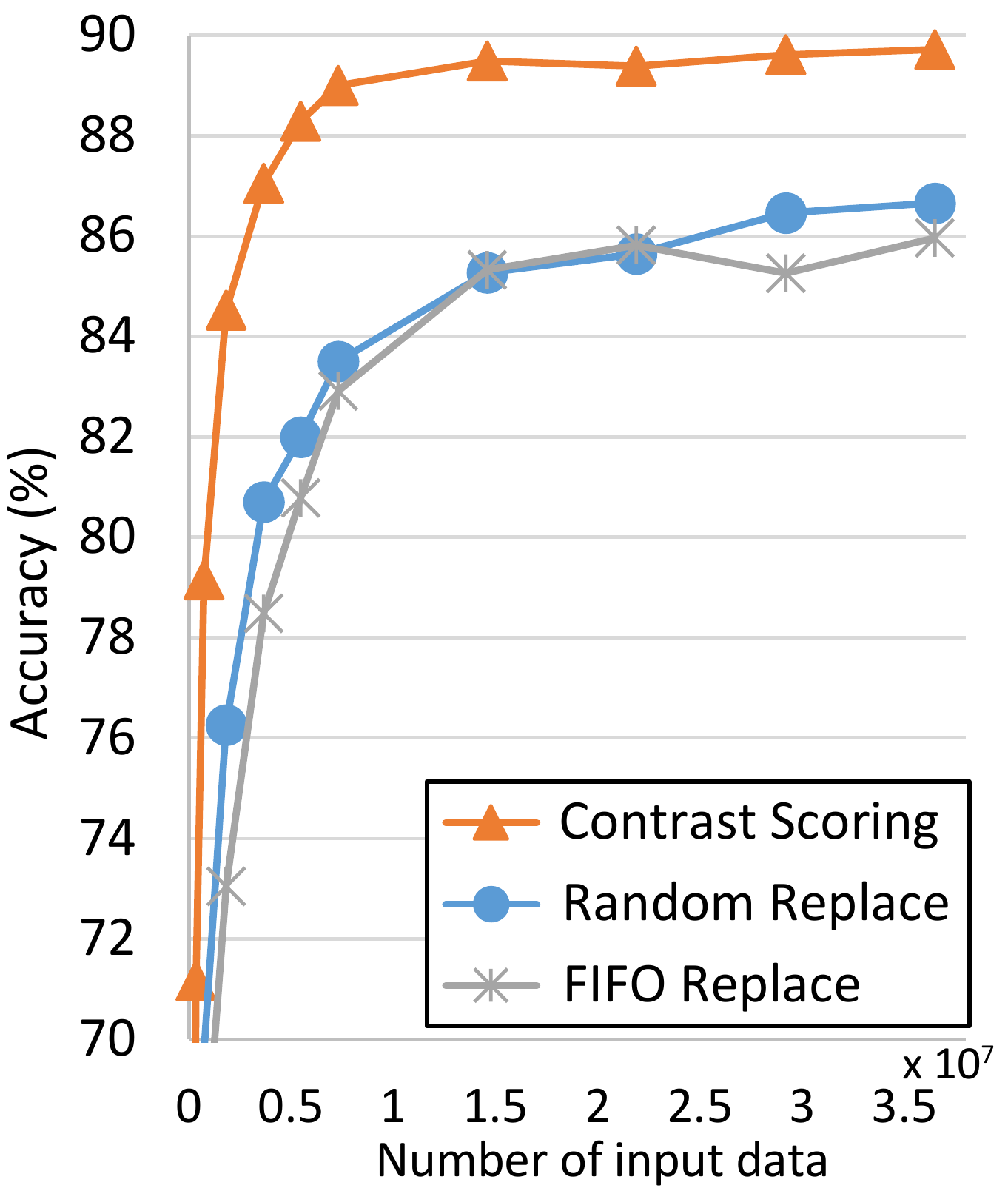}
		\vspace{-16pt}
		\caption{SVHN.}
		\label{fig:}
	\end{subfigure}
		\begin{subfigure}[b]{0.45\linewidth}
		\includegraphics[width=\linewidth]{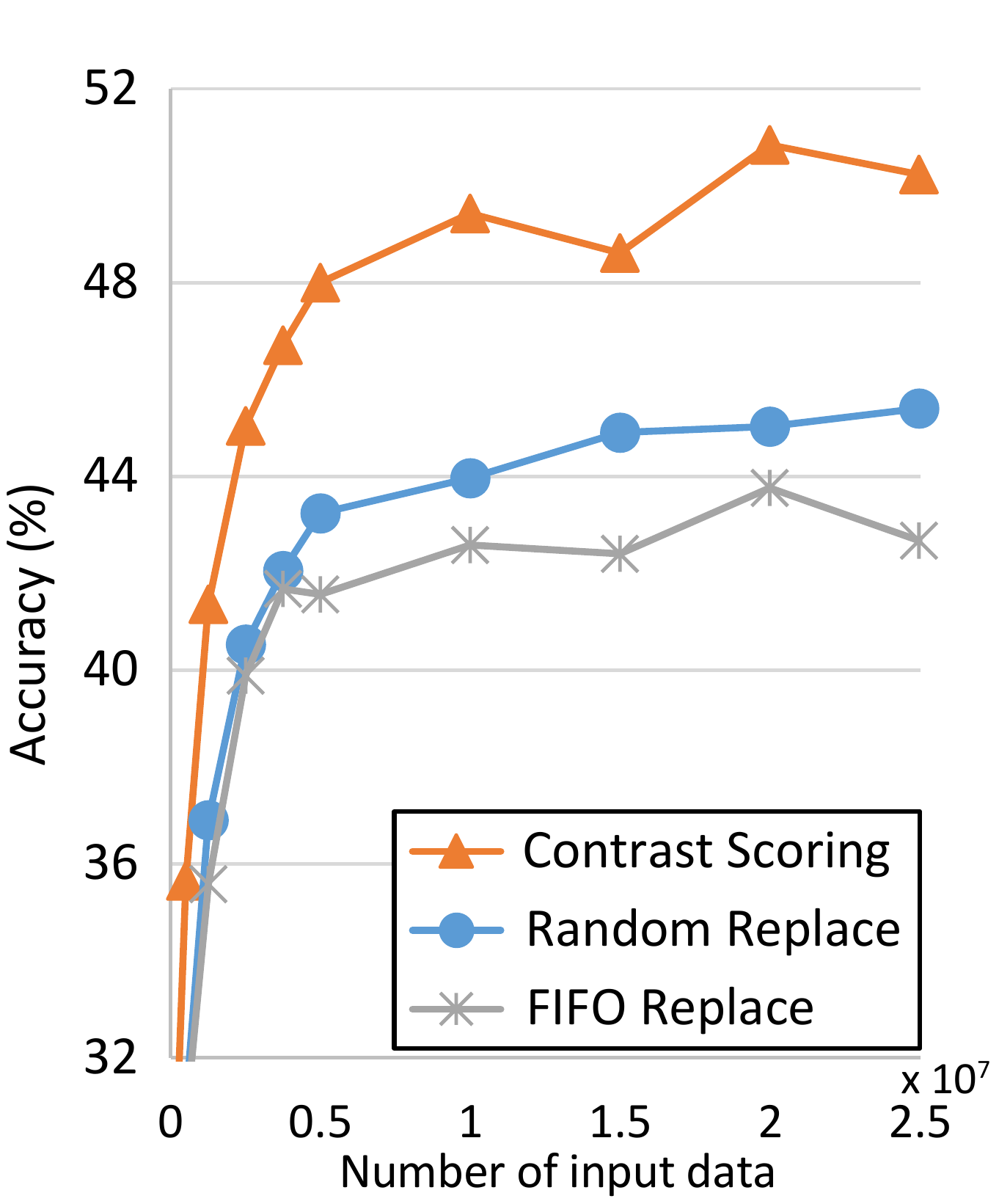}
		\vspace{-16pt}
		\caption{CIFAR-100.}
		\label{fig:}
	\end{subfigure}
	\vspace{-6pt}
	\caption{Learning curve on SVHN and CIFAR-100 datasets.}
	\label{fig:learning_curve_svhn}
	\vspace{-8pt}
\end{figure}

\textbf{Learning Curve on SVHN and CIFAR-100.}
We evaluate the learning curve on the SVHN and CIFAR-100 datasets, and the results are shown in Fig. \ref{fig:learning_curve_svhn}. The learning curve of the proposed approaches substantially outperforms the baselines. 
On the SVHN dataset, 
the proposed approaches achieve 89.71\% accuracy, while the baselines only achieve 86.66\% and 85.96\%, respectively. 
On the CIFAR-100 dataset, the proposed approaches and the baselines achieve 50.22\%, 45.40\%, and 42.68\% top-1 accuracy, respectively.

\subsection{The Impacts of Lazy Scoring.}

\begin{table}[ht]
	\centering
	\caption{Top-1 accuracy, average re-scoring percent, and batch time (relative to that without scoring) on CIFAR-10 with different lazy scoring intervals.}
	\label{tab:lazy_score}
	\setlength\tabcolsep{2.5pt}
    \renewcommand{\arraystretch}{0.8}
    \vspace{-2pt}
    \resizebox{0.85\columnwidth}{!}{
	\begin{tabular}{c|cccccc}
        \toprule
        \begin{tabular}[c]{@{}c@{}}Lazy Scoring \\ Interval\end{tabular} & Disabled & 4                                                       & 20                                                      & 50                                                      & 100                                                     & 200                                                     \\ \midrule
        Accuracy (\%)                                                    & 76.06    & \begin{tabular}[c]{@{}c@{}}77.04\\ (+0.98)\end{tabular} & \begin{tabular}[c]{@{}c@{}}77.18\\ (+1.12)\end{tabular} & \begin{tabular}[c]{@{}c@{}}\textbf{77.23}\\ (\textbf{+1.17})\end{tabular} & \begin{tabular}[c]{@{}c@{}}76.38\\ (+0.32)\end{tabular} & \begin{tabular}[c]{@{}c@{}}74.22\\ (-1.84)\end{tabular} \\ \midrule
        \begin{tabular}[c]{@{}c@{}}Re-scoring \\ Pct. (\%)\end{tabular}  & 100.0    & 21.78                                                   & 4.31                                                    & 1.71                                                    & 0.89                                                    & 0.44                                                    \\ \midrule
        \begin{tabular}[c]{@{}c@{}}Relative \\ Batch Time\end{tabular}   & 1.478    & 1.312                                                   & 1.232                                                   & 1.199                                                   & 1.191                                                   & 1.172                                                   \\ \bottomrule
    \end{tabular}}
    \vspace{-12pt}
\end{table}

We also evaluate the impact of lazy scoring on the accuracy, runtime overhead, and average percent of re-scored data in the buffer in each training iteration.
The model is trained on the CIFAR-10 dataset with buffer size 256 and STC 500.

Lazy scoring effectively reduces the additional computation for scoring during training
and reduces the batch time.
As shown in Table \ref{tab:lazy_score}, when lazy scoring interval $T$ in Eq.(\ref{eq:lazy_score_set}) increases, the average re-scoring percent and the relative batch time (runtime overhead) are effectively reduced. 
When lazy scoring is not used, each training step of our method is 47.8\% slower than the baselines (without scoring). 
When lazy scoring is employed with interval 50, each training step is only about 19.9\% slower than the baselines.
Besides, lazy scoring slightly increases the final accuracy by up to 1.17\%. 
We conjecture that the increased accuracy is because the lazy scoring performs similarly to the momentum encoder in \cite{he2020momentum}. 
The score computed multiple iterations ago serves as a momentum score. This slowly updated score brings more information from the past and benefits the data selection.

\subsection{Improved Accuracy With Different Buffer Sizes}

We evaluate the impact of buffer size on the performance of the proposed approaches.
The model is trained on the CIFAR-10 dataset. The buffer size is in \{8, 32, 128, 256\}. The corresponding learning rate is scaled to \{1, 3, 5, 10\}$\times 10^{-5}$,
roughly following a $\text{learning rate} \propto \sqrt{\text{batch size}}$ scaling scheme.

\begin{table}[!htb]
	\centering
	\caption{Accuracy on CIFAR-10 dataset with different buffer sizes.}
	\label{tab:buffer_size}
	\setlength\tabcolsep{12pt}
	\renewcommand{\arraystretch}{0.8}
	\resizebox{0.8\columnwidth}{!}{
	\begin{tabular}{clc}
		\toprule
		Buffer Size          & \multicolumn{1}{c}{Method} & Accuracy     \\ \midrule
		\multirow{3}{*}{8}   & Contrast Scoring                    & 69.38             \\
		& Random Replace                    & 66.71 (-2.67)  \\
		& FIFO Replace                      & 65.91 (-3.47)  \\ \midrule
		\multirow{3}{*}{32}  & Contrast Scoring                    & 73.26                 \\
		& Random Replace                    & 70.65 (-2.61)  \\
		& FIFO Replace                      & 70.80(-2.46)   \\ \midrule
		\multirow{3}{*}{128} & Contrast Scoring                    & 73.97                  \\
		& Random Replace                    & 71.28 (-2.69)  \\
		& FIFO Replace                      & 70.65 (-3.32)  \\ \midrule
		\multirow{3}{*}{256} & Contrast Scoring                    & \textbf{76.06}                 \\
		& Random Replace                    & 72.75(\textbf{-3.31})  \\
		& FIFO Replace                      & 70.53 (\textbf{-5.53}) \\ 
		\bottomrule
	\end{tabular}}
	\vspace{-6pt}
\end{table}

The proposed approaches consistently outperform the baselines under different buffer sizes.
As shown in Table \ref{tab:buffer_size}, under different buffer sizes, the accuracy by the proposed approaches maintains a clear margin over the baselines. 
Besides, the margin becomes larger as the buffer size increases. This is because a larger buffer size provides the framework a better opportunity to select more informative data, and the proposed approaches can leverage this opportunity to maintain more representative data in the buffer for learning, while the baselines cannot.
Also, all the approaches achieve higher accuracy when the buffer size becomes larger.
This is because a larger buffer size provides a larger batch size, and contrastive learning naturally benefits from a large batch size since it provides more negative samples \cite{chen2020simple}.

\section{Conclusion}
This work aims to enable on-device contrastive learning from input streaming data.
We propose a framework to maintain a small data buffer filled with the most representative data for learning. To achieve the data selection without requiring labels, we propose a data replacement policy by contrast scoring.
Experimental results on multiple datasets show that the proposed approaches achieve superior learning speed and accuracy compared with SOTA baselines.

\noindent
\textbf{Acknowledgement: }
This work used the Extreme Science and Engineering Discovery Environment (XSEDE), which is supported by National Science Foundation grant number ACI-1548562. Specifically, it used the Bridges-2 system, which is supported by NSF award number ACI-1928147, at the Pittsburgh Supercomputing Center (PSC).
This research was supported in part by the University of Pittsburgh Center for Research Computing through the resources provided.

\bibliographystyle{IEEEtran}
\bibliography{IEEEabrv,IEEEexample}

\end{document}